\documentclass[runningheads]{llncs}
\usepackage{graphicx}
\usepackage{subfig}
\usepackage{multirow}
\usepackage{eurosym}
\usepackage{listings}
\usepackage{booktabs} 
\usepackage{multicol}
\usepackage{color}
\usepackage{enumitem}
\usepackage{textcomp}
\usepackage[colorinlistoftodos]{todonotes}
\usepackage{tikz}
\newcommand\copyrighttext{%
  \footnotesize   Copyright © 2020 for this paper by its authors. Use permitted under Creative Commons License Attribution 4.0 International (CC BY 4.0).}
\newcommand\copyrightnotice{%
\begin{tikzpicture}[remember picture,overlay]
\node[anchor=south,yshift=10pt] at (current page.south) {\fbox{\parbox{\dimexpr\textwidth-\fboxsep-\fboxrule\relax}{\copyrighttext}}};
\end{tikzpicture}%
}

\begin{document}

\title{Semantic Entity Enrichment by Leveraging Multilingual Descriptions for Link Prediction}

\author{Genet Asefa Gesese\inst{1,2} \and Mehwish Alam\inst{1,2}
\and Harald Sack\inst{1,2}}
\authorrunning{Genet Asefa Gesese et al.} 
\titlerunning{Link Prediction using Multilingual Entity Descriptions}
\tocauthor{Genet Asefa Gesese, Mehwish Alam and Harald Sack}
\institute{FIZ Karlsruhe -- Leibniz Institute for Information Infrastructure, Germany \\
\and
Karlsruhe Institute of Technology, Institute AIFB, Germany \\
\email{firstname.lastname@kit.edu}}

\maketitle             
\begin{abstract}
Most Knowledge Graphs (KGs) contain textual descriptions of entities in various natural languages. These descriptions of entities provide valuable information that may not be explicitly represented in the structured part of the KG. Based on this fact, some link prediction methods which make use of the information presented in the textual descriptions of entities have been proposed to learn representations of (monolingual) KGs. However, these methods use entity descriptions in only one language and ignore the fact that descriptions given in different languages may provide complementary information and thereby also additional semantics. In this position paper, the problem of effectively leveraging multilingual entity descriptions for the purpose of link prediction in KGs will be discussed along with potential solutions to the problem.
\end{abstract}

\copyrightnotice
\section{Introduction}\label{intro}
Knowledge Graphs (KGs) such as Freebase \cite{bollacker2008freebase}, DBpedia \cite{lehmann2015dbpedia}, and Wikidata \cite{vrandevcic2014wikidata} have been created in order to share linked data which describes entities and the relationships between them. The availability of these various cross-domain KGs has sparked interest in undertaking research directions such as KG completion using tasks like link prediction. Hence, different Knowledge Graph Embedding (KGE) approaches, which map KGs to a low dimensional vector space, based on a link prediction task have been published. Link prediction is widely used because in the Open World Assumption the knowledge explicitly represented in a KG is never complete, there are always missing facts which can be predicted using link prediction.

DistMult \cite{yang2014embedding} and ConvE \cite{dettmers2018convolutional} are among those KGE models which are trained on a link prediction task but without making use of the textual descriptions of entities. On the other hand, there are some models such as DKRL \cite{xie2016representation} which leverage the textual descriptions of entities for the link prediction task on (monolingual) datasets like FB15K \cite{bordes2013translating} and FB15K-237 \cite{toutanova-chen-2015-observed} and have demonstrated that using the textual descriptions enhance the embeddings of entities \cite{gesese2019survey}. However, despite the fact that the entities in these KGs have multilingual entity descriptions, all the existing models which use descriptions of entities focus on using descriptions written in only one natural language. 

\begin{figure}
    \centering
    \includegraphics[scale=0.4]{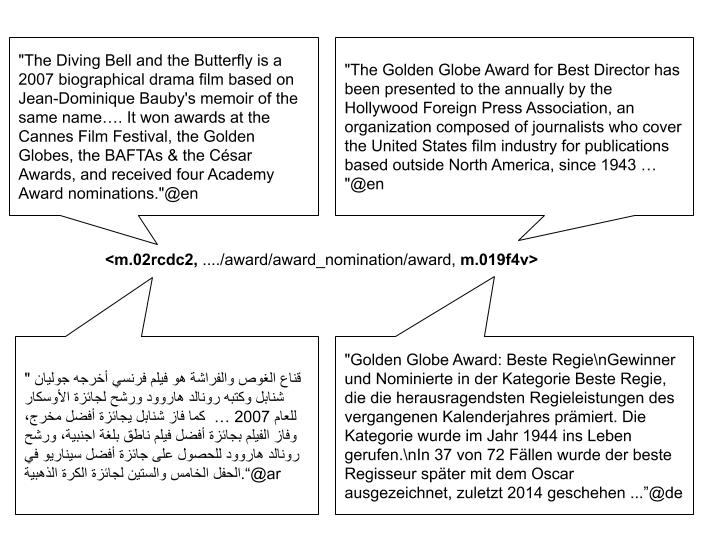}
    \caption{A triple in FB15K with multilingual descriptions of the head and tail entities from Freebase.}
    \label{fig:triple}
\end{figure}

In most of the popular KGs, a single entity can have descriptions in two or more languages where the contents of the descriptions are different. This fact is demonstrated in Figure \ref{fig:triple} using, as an example, a triple from FB15K with some of the descriptions of its head and tail entities extracted from Freebase. In this example, it can be seen that, for both the head (`m.02rcdc2') and tail (`m.019f4v') entities, the description provided in one language contains information that is not available in the description given in the other language. Hence, a KGE model which uses entity descriptions only in one language discards the extra information provided in the descriptions in other languages.

\section{Related Work}\label{rel}

Few attempts have been made to combine the structured part of KGs with entity descriptions to learn KGE models. Among these models, DKRL \cite{xie2016representation},  MKBE \cite{pezeshkpour-etal-2018-embedding}, and Jointly \cite{xu2017knowledge} use neural network encoders (either CNN or LSTM) to represent entity descriptions. The other models are SSP \cite{xiao2017ssp} and LiteralE \cite{kristiadi2019incorporating} which rely on document embedding approaches to get representations for entity descriptions. In DKRL, CNN is used to encode entity descriptions using word embeddings as an input. 
MKBE, same as in DKRL, uses CNN to encode textual descriptions of entities. The descriptions that are used in both DKRL and MKBE are provided in only one language (English).
    
Jointly \cite{xu2017knowledge} is a KGE method which combines structural and textual encoding as in DKRL but using (attentive) LSTM encoder instead of CNN. In this approach, the embedding of a word is initialized by taking the average of the embeddings of the entities whose description include this word. Initialising in this way does not work well for multilingual descriptions because it is not capable of capturing words which are from different languages but semantically similar and are not linked to the same set of entities.

SSP is another KGE approach which jointly learns from structured information and entity descriptions. This method adopts the Non-negative Matrix Factorization (NMF) topic model to generate a representation for an entity based on its description, i.e., treating each entity description as a document and taking the topic distribution of the document as the representation of the corresponding entity. However, the approach would not perform well with multilingual entity descriptions since the adopted topic model does not deal with multilinguality.
In LiteralE, entity descriptions are represented using a document embedding technique proposed in \cite{le2014distributed}. This document embedding technique works by first mapping the whole document (i.e., entity description) and also every word present in the document into corresponding unique vectors and then taking the average or concatenation of the paragraph vector and word vectors so as to predict the next word in a context.

\section{Proposed Methodology} \label{method}
In order to address the issues with the existing KGE models in using multilingual descriptions, this study provides the following insights into the potential solutions.  

\subsection{Applying Language Translators}
The straight forward way to incorporate multilingual entity descriptions in the existing neural network encoder based KGE models (i.e., DKRL, MKBE, and Jointly) is first to convert all the descriptions into one language (English) with a language translator and then to pass as inputs to the encoder pre-trained embeddings of the words present in the descriptions. The pre-trained word embeddings can be obtained from any monolingual word embedding model. The main challenge with this method is the errors that occur during machine translation (converting multilingual descriptions into one language) will be propagated to the encoder. One way to address this issue is to use the multilingual word embeddings instead of applying machine translation.

\subsection{Using Multilingual Word Embeddings}
KDCoE \cite{chen2018co} is a KGE approach which leverages a weakly aligned multilingual KG for semi-supervised cross-lingual learning using descriptions of entities. With this approach, the authors have demonstrated that a very good performance can be achieved for an entity alignment task by using an Attentive Gated Recurrent Unit encoder (AGRU) to encode multilingual descriptions with multilingual word embeddings as inputs. The multilingual embedding model used is a cross-lingual Bilbowa word embedding \cite{gouws2015bilbowa} trained on the cross-lingual parallel corpora Europarl v7 \cite{koehn2005europarl} and monolingual corpora of Wikipedia dump. The results from KDCoE show that multilingual text encoders can benefit from multilingual word embeddings. It would also be interesting to adopt the same approach, which is used to encode multilingual entity descriptions for cross-lingual entity alignment task in KDCoE, for a link prediction task on different monolingual datasets such as FB15K and FB15K-237. This approach allows to capture as much information as possible from entity descriptions present in multiple languages. 

\begin{figure}
    \centering
    \includegraphics[scale=0.4]{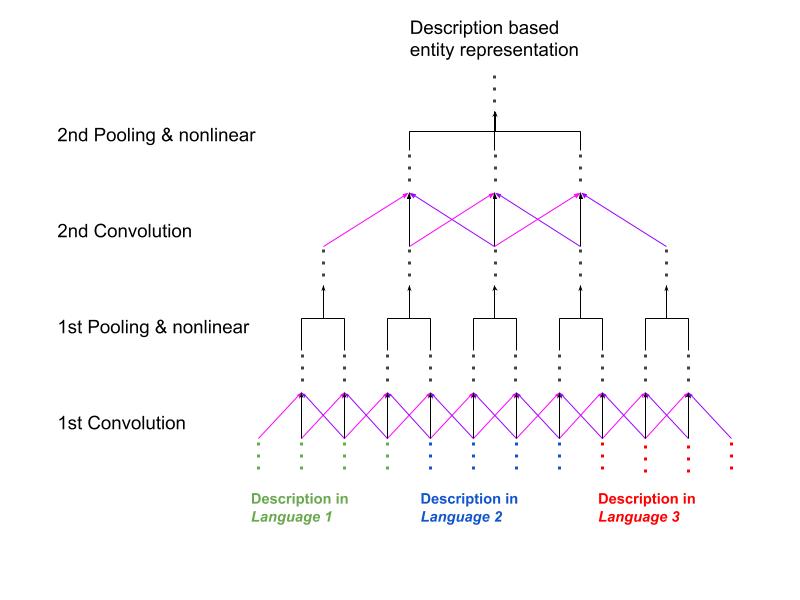}
    \caption{Passing pre-trained multilingual word embeddings to a CNN encoder which is adopted from DKRL \cite{xie2016representation}, in order to encode multilingual entity descriptions.}
    \label{fig:cnn}
\end{figure}

Furthermore, the existing models such as DKRL and Jointly can be improved by leveraging multilingual entity descriptions by passing as inputs to the encoders the embeddings of the words in the descriptions obtained by a multilingual word embedding model like MUSE \footnote{https://github.com/facebookresearch/MUSE}. For instance, Figure \ref{fig:cnn} shows how the CNN encoder part of DKRL can be modified to take pre-trained word embeddings from multilingual descriptions as inputs.

\section{Conclusion}
In this position paper, the problem of leveraging multilingual entity descriptions for link prediction task on KGs is discussed. As mentioned in Section \ref{intro} and Section \ref{rel}, the available link prediction models on monolingual datasets such as FB15K-237 use only monolingual entity descriptions and ignore the fact that the descriptions in other languages may contain additional semantics. Thus, in this study, some insights into potential solutions to this problem are provided. These solutions enable the existing link prediction models to leverage multilingual entity descriptions. The solutions proposed in this study for link prediction task can also be adopted for other KG completion tasks such as triple classification and entity classification. In order to come up with an even better solution to the problem, conducting detailed analysis on the nature and quality of the multilingual entity descriptions available in different KGs such as DBpedia, Wikidata, and Freebase would be beneficial.

\newpage
\bibliographystyle{splncs04}
\bibliography{biblio}

\end{document}